\documentclass[runningheads]{llncs}
\usepackage{graphicx}
\usepackage[thinc]{esdiff}
\usepackage{amsmath}
\usepackage{todonotes}
\usepackage{multirow}
\usepackage{color, colortbl,booktabs}
\usepackage[hidelinks]{hyperref}
\definecolor{Gray}{gray}{0.9}
\begin{document}
\title{FairFlow: An Automated Approach to Model-based Counterfactual Data Augmentation For NLP}

\titlerunning{FairFlow}
%
\author{Ewoenam Kwaku Tokpo \and Toon Calders}
\authorrunning{E. Tokpo et al.}
%
\institute{University of Antwerp, Antwerp, Belgium
\email{\{ewoenamkwaku.tokpo,toon.calders\}@uantwerpen.be}}
\maketitle              

\begin{abstract}
Despite the evolution of language models, they continue to portray harmful societal biases and stereotypes inadvertently learned from training data. These inherent biases often result in detrimental effects in various applications.
Counterfactual Data Augmentation (CDA), which seeks to balance demographic attributes in training data, has been a widely adopted approach to mitigate bias in natural language processing. 
However, many existing CDA approaches rely on word substitution techniques using manually compiled word-pair dictionaries. These techniques often lead to out-of-context substitutions, resulting in potential quality issues. The advancement of model-based techniques, on the other hand, has been challenged by the need for parallel training data. Works in this area resort to manually generated parallel data that are expensive to collect and are consequently limited in scale.
This paper proposes FairFlow, an automated approach to generating parallel data for training counterfactual text generator models that limits the need for human intervention. 
Furthermore, we show that FairFlow significantly overcomes the limitations of dictionary-based word-substitution approaches whilst maintaining good performance.

\keywords{ Natural language processing \and Bias mitigation \and Counterfactual Data Augmentation}

\end{abstract}
\section{Introduction}
\label{sec: intro}
Despite their growing popularity and unprecedented performance in various application domains, language models (LMs) continue to be plagued with issues of harmful societal biases and stereotypes that have been shown to have detrimental social effects \cite{dastin2022amazon}. 
The biggest contributing factor is the encapsulation of societal biases in everyday language, as is well-documented \cite{beukeboom2017linguistic,porter2016inferring,fiedler2011social}. LMs heavily rely on such textual data, now digitalized on various online outlets, as training data, causing them to mirror these biases \cite{wolf2017we}. 

In Natural Language Processing (NLP), similar to many machine learning domains, bias mitigation generally occurs at three intervention avenues: the training data, the learning procedure, or the model output  \cite{lohia2019bias}.
Since model bias traces its roots to the training data, mitigating bias at the training data level has proven very effective \cite{dixon2018measuring,de2019bias}. 
One such approach, Counterfactual Data Augmentation (CDA) \cite{dattagender}, seeks to remove spurious correlations between attributes in the training data by evening out the distribution of words that characterize demographic attributes in the context of neutral words that should ideally not be demographically aligned. Specifically, explicit attribute-defining words are replaced with their counterfactual equivalents from complementary demographic groups for every text instance. 
To illustrate this with an example, an instance of ``\textit{She is a nurse}'' will be augmented with ``\textit{He is a nurse}'' in the case of mitigating gender bias. This follows the intuition that in an ideal dataset, the association between gender attributes and target attributes like professions will be even for different gender groups. 

Key works, such as \cite{zhao-etal-2018-gender,lu2020gender,zmigrod-etal-2019-counterfactual}, introducing CDA as a bias mitigation technique adopt a word substitution approach based on dictionaries. 
These word substitution methods are prone to grammatical incoherence because of out-of-context substitutions and omitted word pairs. 
Because dictionary compilations are often incomplete \cite{dinan2020queens}, a direct word-substitution approach will not generalize to omitted words.
Take for instance (\textit{\textbf{Bachelor} and \textbf{Masters} degree} v. \textit{\textbf{Spinster} and \textbf{Mistresses} degree}) and (\textit{she taught \textbf{herself}} v. \textit{he taught \textbf{herself}}) which were common issues we observed with some methods.
Additionally, the dictionaries are manually compiled, which not only incurs potential costs but manually compiling counterfactual word pairs for certain demographics may be intrinsically challenging. 

Although generative language models like GPT-related models \cite{radford2018improving} have surged in popularity, their adoption for CDA has been limited due to the relative unavailability of parallel data needed for training.  As such, model-based solutions resort to manually compiling parallel training data, a process that is both costly and constrained. This challenge is exacerbated by the fact that training models on limited parallel data can impair performance \cite{zoph2016transfer}.
Although large conversational models like ChatGPT generate good counterfactuals in a zero-shot setting, they are not efficient in low-resource environments. In this work, we focus on low-resource/resource-efficient techniques that can be deployed in low-resource environments.

\begin{figure}[tbh]
    \centering
    \includegraphics[width=1\linewidth]{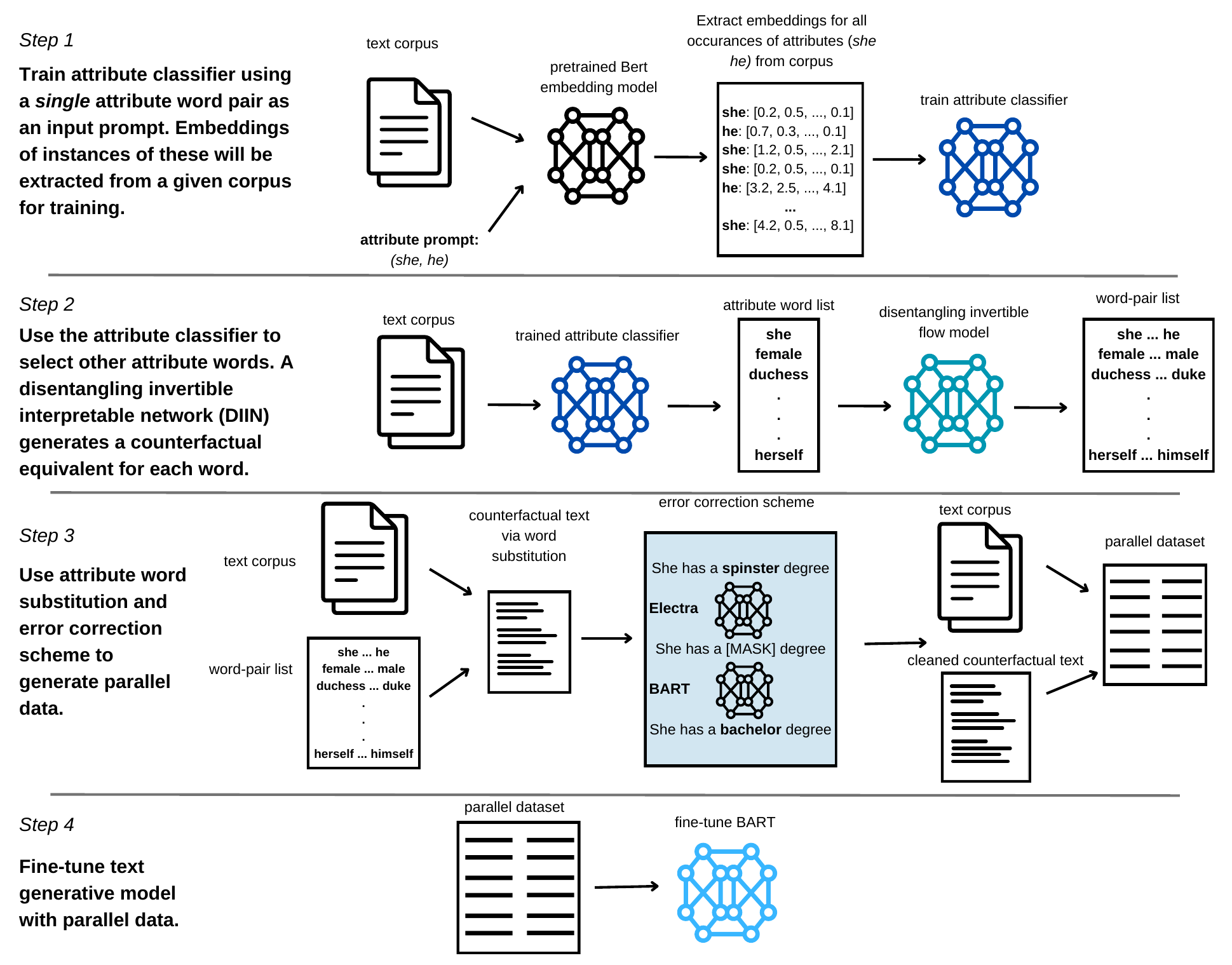}
    \caption{An end-to-end description of Fairflow, described in four steps: 1) train a classifier to identify attribute words from a corpus; 2) generate counterfactual equivalents for attribute words using an invertible generative flow model; 3) use a word substitution scheme and our proposed error-correction scheme to make the parallel text more fluent and realistic; 4) fine-tune a generative model with the generated parallel data.}
    \label{fig: architecture}
\end{figure}

The primary contribution of this paper is to explore an automated approach to generate parallel training data for a given demographic axis that requires minimal human intervention. 
Our approach takes from a user a prompt -- in the form of a single word-pair --  that describes a demographic axis. This pair is subsequently used to model a demographic subspace from which other words that define the demographic attribute can be sampled from a given corpus of text. Using an invertible flow-based model \cite{dinh2014nice}, counterfactual words are generated for sampled words.
Thereafter, an error correction approach is used in tandem with direct word substitution to generate parallel data to fine-tune a generative language model to generate counterfactual texts. We call our approach and the resultant counterfactual text generation model \textit{FairFlow}.
This entire process is simply depicted in a four-step process in Fig.~\ref{fig: architecture}.
As opposed to existing works, which will be discussed in Section ~\ref{sec: related_lit}, FairFlow does not rely on human-generated parallel data for training and eliminates the need for manually compiled word-pair dictionaries. 

In summary, this paper explores and proposes techniques to develop a robust model-based counterfactual generator in the absence of parallel training data. Key contributions include:
\begin{enumerate}
    \item An automated approach to compiling dictionaries of word pairs that only requires a user to input a word-pair prompt that describes a demographic axis. 
    \item We proposed an error correction approach to generate parallel data from dictionary word substitutions.
    \item We train a counterfactual model using our generated parallel data and show that the error correction approach not only improves the grammatical composition of the model but also improves the generalization of the model.
\end{enumerate}

We make our implementation code and materials for FairFlow available\footnote{\href{https://github.com/EwoeT/FairFlow}{https://github.com/EwoeT/FairFlow}}.

\section{Background and Related Literature}
\label{sec: related_lit}
Early works on CDA used simple rule-based word-substitution approaches for counterfactual data augmentation. Specifically, they created dictionaries of attribute word pairs and used matching rules to swap words \cite{de2019bias}. 
Later works began to incorporate grammatical information like part-of-speech tags to swap attribute words \cite{zhao-etal-2018-gender}. 
In the absence of interventions for named entities, Lu et al. \cite{lu2020gender} do not augment sentences or text instances containing proper nouns, and named entities as generating counterfactuals without proper name interventions could result in semantically incorrect sentences.
Zhao et al. \cite{zhao-etal-2018-gender} circumvented this by anonymizing named entities by replacing them with special tokens.
Lamenting on the aforementioned lack of parallel corpus for training neural models, Zmigrod et al. \cite{zmigrod-etal-2019-counterfactual} used a series of unsupervised techniques such as dependency trees, lemmata, part-of-speech tags, and morpho-syntactic tags for counterfactual generation. 
Hall-Maudsley et al. \cite{hall-maudslay-etal-2019-name} improve on Zmigrod et al. by incorporating a names intervention method to resolve the challenges of generating counterfactuals for named entities. They achieve this using a bipartite graph to match first names.

Because the aforementioned techniques rely on dictionary word replacement techniques and ignore the context of the text, they are prone to generating ungrammatical texts.
Additionally, the inability of these techniques to resolve out-of-dictionary words not only preserves certain attribute correlations but also introduces errors.
We illustrate two instances of such limitations using the word substitution approach by Hall-Maudsley et al. on the Bias-in-bios dataset \cite{de2019bias}; 1) \emph{``Memory received her \textbf{Bachelor} and \textbf{Masters} of Accountancy...''}  produces \emph{``Memory received his \textbf{Spinster} and \textbf{Mistresses} of Accountancy...''} due to the polysemous nature of \emph{bachelor} and \emph{master};
2) \emph{"Laura discovered her passion for programming after teaching \textbf{herself} some Python..."}, is transformed into 
\emph{"Anthony discovered his passion for programming after teaching \textbf{herself} some Python..."} as the gender pronouns \emph{herself} and \emph{himself} are excluded from the dictionary compiled by Hall-Maudsley et al.

More recently, sequence-to-sequence model-based approaches to counterfactual generation have been proposed \cite{wu2021polyjuice,qian2022perturbation}. Wu et al. \cite{wu2021polyjuice} propose Polyjuice, a generative counterfactual model for diverse use cases like counterfactual explanations. They generate parallel data by pairing naturally occurring sentences in a corpus based on edit distances. Although effective for explanations, such an approach is not applicable for bias mitigation as attribute words, in the case of the latter, have to be specifically defined and replaced.
Specifically for bias mitigation, Qian et al. \cite{qian2022perturbation} introduce the \textit{perturber}, which is a Bart\cite{lewis-etal-2020-Bart} model fine-tuned on a human-generated parallel text. 
However, their approach only generates counterfactuals for specific user-defined entities in a text. eg. 
\textit{original:``Torii chose to remain behind, pledging that he and his \textbf{men} would fight...'', rewrite:``Tara chose to remain behind, pledging that she and her \textbf{men} would fight ...''}.
As earlier stated, such manually compiled datasets are expensive and are only available on small scales, which can degrade performance \cite{zoph2016transfer}.
Additionally, similar manual efforts must be solicited for every language domain for which counterfactuals have to be generated.
As opposed to existing works, the main advantage of our work is the non-reliance on human-generated parallel data and word lists.

\section{Approach}
\label{sec: approach}
Our entire approach can be summarized in four steps as illustrated in Fig.~\ref{fig: architecture}.
The process commences with training a classifier to detect attribute words in a corpus, after which counterfactuals for these attribute words are generated using an invertible flow model. Parallel data is thereafter created by using a combination of word substitution and an error-correction scheme. Finally, a generative model is fine-tuned using the generated parallel data. We expound on these steps in the following subsections.

\subsection{Attribute classifier training}
To select a list of words that characterize a given demographic axis, e.g. gender, we first train an attribute classifier that approximates the attribute subspace. 
To do this, the user first inputs a prompt in the form of a single pair of words that describes a given demographic axis, e.g., (she, he) in the case of gender.
Using a pretrained contextualized word embedding model, contextualized word representations are generated for each appearance of the input words within a given text corpus  — we take \textit{BERT-base-uncased} \cite{devlin2018bert} as our choice of representation model. 
These embeddings are used to train a classifier to approximate the demographic subspace.
Formally, consider the word-pair \((x_a, x_b)\) that define a demographic axis, we obtain two sets \(Z_a=\{z_{a_1}, z_{a_2}, ..., z_{a_n}\}\) and \(Z_b=\{z_{b_1}, z_{b_2}, ..., z_{b_n}\}\) where \(z_{a_i} \in R^{d} \) and \(z_{b_i} \in R^{d} \) are context-specific vector representations of instances of \(x_a, x_b\) respectively, generated from a text corpus \(V\) by a pretrained embedding model \(E\); so that \(E(x_i, c_i)=z_{i}\) if \(x_i\) is an instance of a word \(x\) and \(c_i\) is its context.
We estimate the demographic subspace by training a classifier \(H\) to maximizing the objective \(\sum\limits_{z_i \in \{Z_a \cup Z_b\}} log (P(y|z_i))\), where \(y=\{a,b\}\) is the class label of \(z_i\). \(H\) is parameterized as a feed-forward neural network with one hidden layer and Gelu non-linear activation.

\subsection{Generating word-pair list}
\subsubsection{Selecting attribute words.}
Given a demographic subspace, we select all words that lie within the attribute-defining regions of the subspace. This process is formally described as follows.
Given our initial corpus \(V\), we select words \(x_i\in V\) based on the criterion \(P(y|E(x_i, c_i);\Theta_H)>\phi\)   where \(\Theta_H\) represents the parameters that define \(H\) and \(\phi\) is a predefined threshold. \(Z_a\) is thus expanded to include all words that have at least an instance satisfying \(P(y=a|E(x_i, c_i);\Theta_H)>\phi\) and \(Z_b\) to include all words with at least an instance satisfying \(P(y=b|E(x_i, c_i);\Theta_H)>\phi\). Although some neutral words may be included in these sets, they do not produce any counterfactual equivalent in the next stage, hence making no difference.

\subsubsection{Generating counterfactual word-pairs with DIIN.}
The first step in generating counterfactual equivalents for the set of words \(Z_a\) and \(Z_b\) is to define a transformation \(T\) from the original embedding space into an ``interpretable'' space where an embedding is factorizable into independent components. We train $T$ to constrain attribute information \textit{only} to the first $k$ dimensions (we will collectively refer to these dimensions as $K$) of a word in the interpretable space. By so doing, $K$ can be swapped to alter the attribute (eg. gender) of the word.
We implement $T$ using a flow-based generative model \cite{kobyzev2020normalizing,papamakarios2021normalizing,dinh2014nice}; specifically, we use the disentangling invertible interpretation network (DIIN) architecture by Esser et al.  \cite{esser2020disentangling}.

Formally, given the contextualized representation \(z\) of a word \(x\), the goal is to learn a transformation \(T\) that maps the original representation  \(z \in R^d\) to an interpretable representation \(\tilde{z} \in R^d\) s.t. \(T(z)=\tilde{z}\).
The interpretable representation \(\tilde{z}\) is sampled from a base distribution \(\tilde{z} \sim p_{\tilde{\mathcal{Z}}}(\tilde{z})\) -- a standard Gaussian distribution in this case.
Using the change of variable theorem, \(T\) is learned by maximizing the log-likelihood 
\begin{equation}
    \log(p_{\mathcal{Z}}(z)) = \log(p_{\tilde{\mathcal{Z}}}(T(z))) + \log(|\det (\diffp{T(z)}{z})|)
\end{equation}  

To constrain attribute information only to $K$, we pair embeddings of words that have the same attribute \(F\) and train \(T\) to generate similar values for both embeddings in their first \(k\) dimensions in the interpretable space.
Mathematically, Given a pair of embeddings \( (z_{a_1}, z_{a_2}) \) that belong to the same demographic group such that  \(F_{z_{a_1}} = F_{z_{a_2}}\), the objective is achieved by minimizing the loss function:
\begin{equation}
\begin{split}
    \mathcal{L}(z_{a_1},z_{a_1}|F) = ||T(z_{a_1})_D||^2 -  \log(\det (T(z_{a_1}))) \\
    + ||T(z_{a_2})_{(D \setminus K)}||^2 - \log(\det (T(z_{a_2})))\\  + \frac{||T(z_{a_2})_K - \sigma T(z_{a_1})_K||^2}{1-\sigma^2} 
\end{split}
\end{equation}
where \(D\) is a term to collectively refer to all $d$ components of the embedding.  \(\sigma\in (0,1)\) is a positive correlation factor that determines the strength of the correlation between \(z_{{a_2}_K}\) and \(z_{{a_1}_K}\).
We also use the dimensionality estimation approach of Esser et al. to estimate the dimensionality of $K$.

\begin{figure}
    \centering
    \includegraphics[clip, trim={7cm 2cm 7cm 2cm},width=0.75\linewidth]{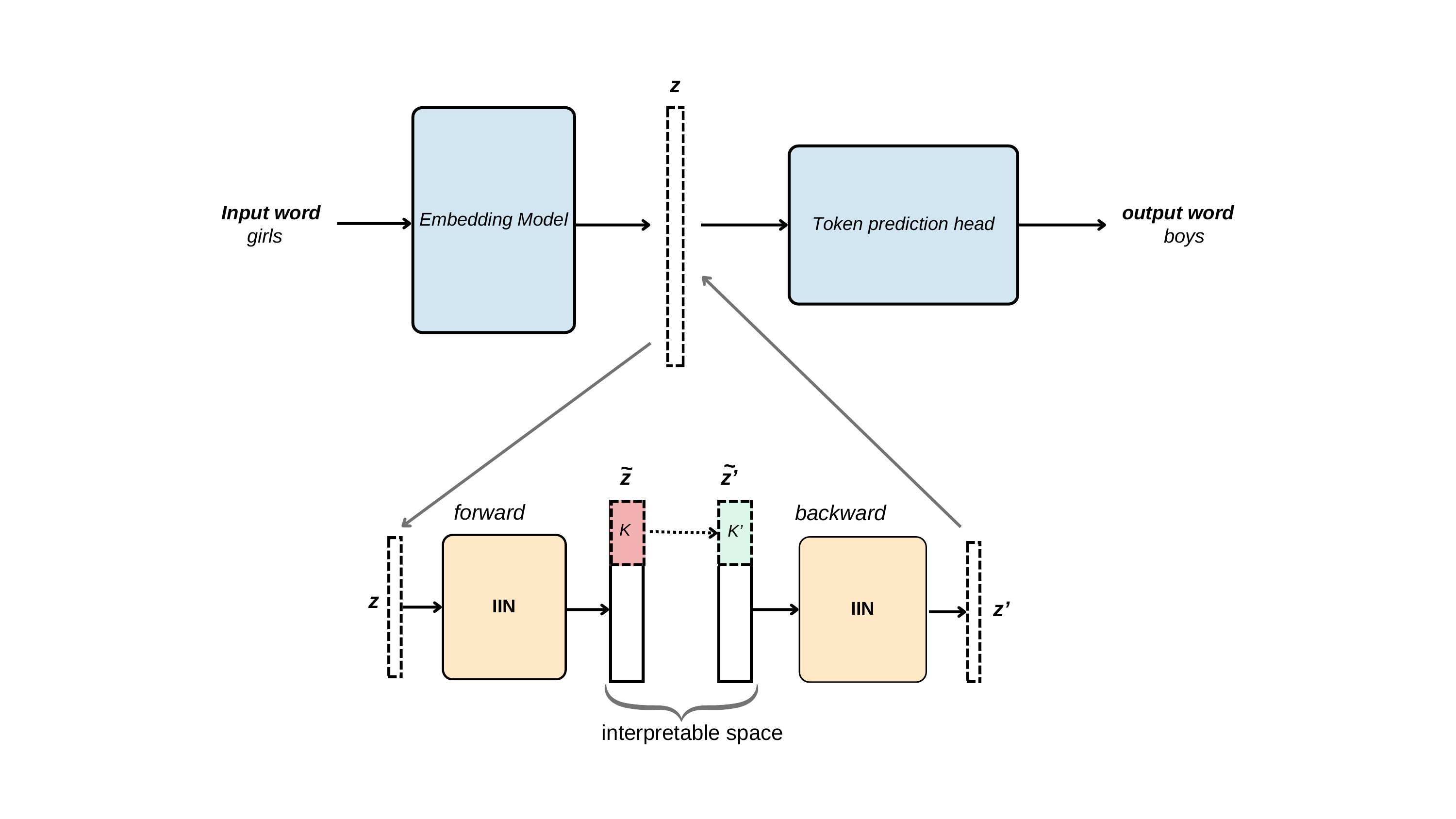}
    \caption{Counterfactual word generation using an invertible interpretation flow network IIN.}
    \label{fig:DIIN}
\end{figure}

\begin{figure}
    \centering
    \includegraphics[clip, trim={2cm 2cm 2cm 2cm},width=1\linewidth]{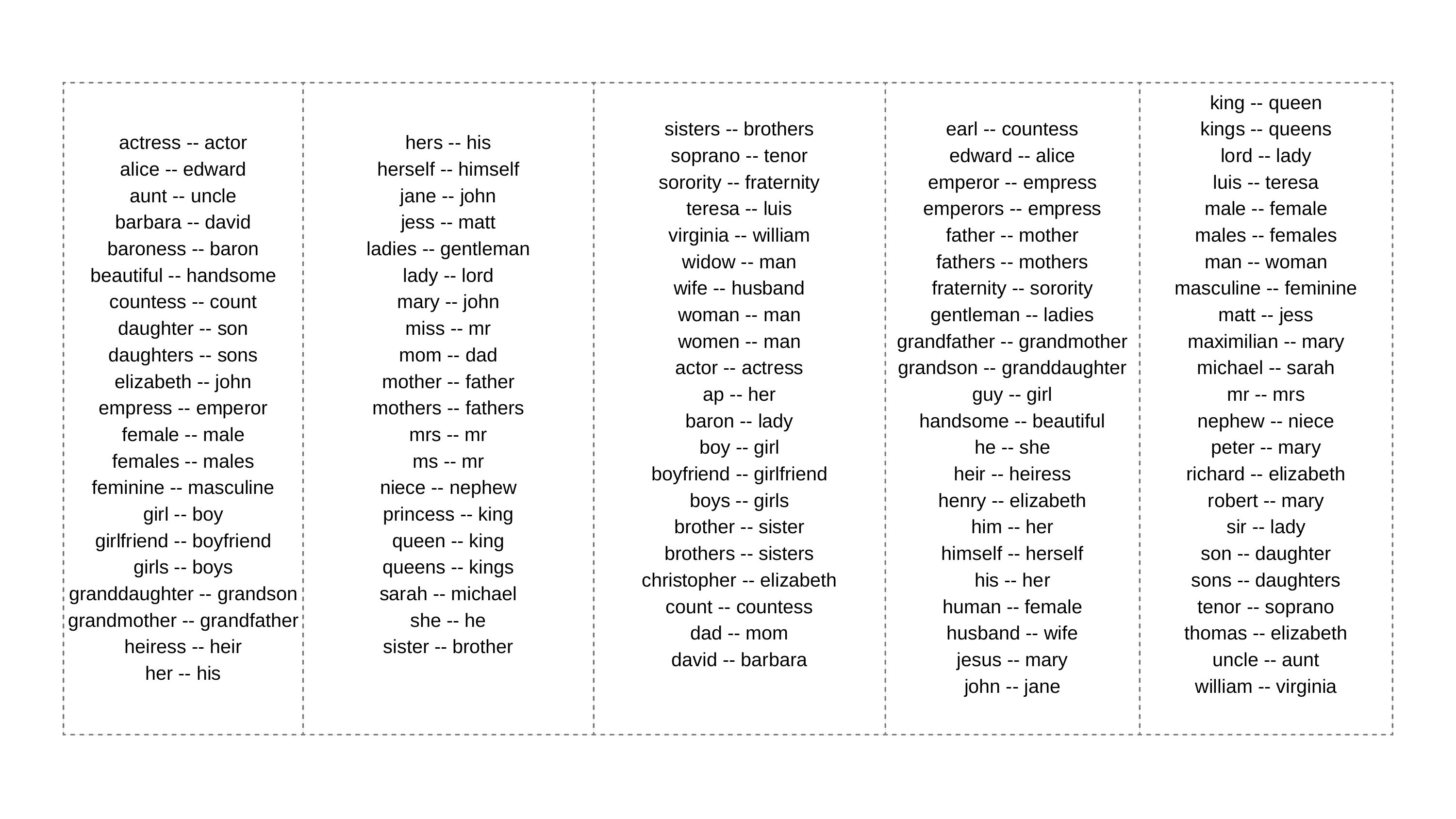}
    \caption{An autmatically compiled dictionary using the input prompt \{\textit{"she"}, \textit{"he"}\}. Words are discovered using the attribute classifier, and the counterfactuals are generated using the disentangling invertible interpretation network.}
    \label{fig:dictionary_samples}
\end{figure}

Once our invertible flow model has been trained to constrain \(F\) to the first \(k\) dimensions of \(\tilde{z}\) (in the interpretable space), 
we replace \(z_{{a_i}_K}\) which is the first \(k\) dimensions of \(\tilde{z}_{a_i}\) with \(K_b^{\prime}\); such that \(z_{{a_i}_K}\rightarrow K_b^{\prime} \), where \(K_b^{\prime} = \frac{1}{N} \sum \limits_{i=0}^N z_{{b_i}_K}\) is the average of the first \(k\) dimensions of the complementary demographic group.
This process is depicted in Fig.~\ref{fig:DIIN}. We use a majority voting scheme to then select the most frequent equivalent generated for each word.
An output example of this process obtained using a \{\textit{"she"}, \textit{"he"}\} prompt is shown in Fig.~\ref{fig:dictionary_samples}.
We then extend this list using the names intervention approach of Hall-Maudsley et al. to generate counterfactuals for names. 

\subsection{Error correction}
\label{subsec: error_correction}
With the word pairs generated from the previous phase, we use the word substitution approach of Hall-Maudsley et al. to build a base corpus. To transform this base corpus into fluent and realistic text labels for our parallel training data, we proposed an error correction scheme which we describe below in two steps.

\subsubsection{Erratic token detection.}
The idea here is to detect and mask tokens that have a low probability of appearing in the context of a given text; following $t_i = t_{<mask>}$ if $P(t_i|T \setminus t_i)<\theta$, where $T$ is the sequence of tokens, $t_i$ is the $i$th token in $T$, and $\theta$ is a predefined threshold value.
We define the resulting masked text as $T_\Pi$.
This is achieved using a pretrained Electra model \cite{clark2020electra}. Electra is an LM pretrained using a text corruption scheme -- text instances are corrupted by randomly replacing a number of tokens with plausible alternatives from BERT. Electra is then trained to predict which tokens are real and fictitious.

Since the use of wordpiece tokenization causes issues (as a word can be broken down into multiple subtokens) if a subtoken is selected for masking, we replace the entire sequence of associated subtokens with a {\tt<mask>} token.
For instance, \emph{``The men are duchesses''}, in a wordpiece tokenization could be decomposed to \emph{[``The'', ``men'', ``are'', ``duchess'', ``\#\#es'']}, Consequently, when \emph{``duchess''} is identified as an erratic token, the masking scheme replaces the entire subsequence \emph{[``duchess'', ``\#\#es'']}, thereby, generating \emph{``The men are {\tt<mask>}''}.

\subsubsection{Text insertion with BART.}
Having obtained our masked intermediary texts, we generate plausible token replacements for each masked token. Since a {\tt<mask>} token could correspond to multiple subword tokens, the replacement generator should be capable of generating multiple tokens for a single {\tt<mask>} instance, making it suitable to use a generative model -- pretrained BART \cite{lewis-etal-2020-Bart} -- to predict these replacement tokens. 
Because Masked Language Modeling is one of BART's pretraining objectives, we can utilize it in its pretrained form without the need for finetuning.
Given $T_\Pi$ from the previous step, the BART model tries to predict the correct infilling \(x\) using the context of $T_\Pi$.

\subsection{Training the generative model}
The final stage of the approach is to fine-tune a BART model using the parallel data obtained from the previous steps. 
The BART generator takes the original text as input and is trained to autoregressively generate the counterfactual of the source text using the corresponding parallel counterfactual texts as labels in a teacher-forcing manner \cite{williamsteacher}. We formulate this as:
\begin{equation}
 \mathcal{L}_{generator} = -\sum\limits_{t=1 }^klogP(y_t|Y_{<t}, X)   
\end{equation}
Where  $X$ and $Y$ are the source and target texts, respectively, $y_t \in Y$ is the $t^{th}$ token in the target text, and $Y_{<t}$ refers to all tokens in $Y$ preceding $y_t$.

\section{Experimental set-up}
This section describes key implementation details of our work and the evaluation framework. 
We specifically evaluate gender bias in the binary sense within the English language domain. 

\subsection{Training set-up}
The main corpus for training the attribute classifier and the disentangling invertible flow model comprises Wikipedia articles via Wikimedia dumps\footnote{https://dumps.wikimedia.org}.

\subsection{Evaluation datasets}

For the appraisal of our model, we used the datasets discussed below. These datasets, upon which various CDA interventions were applied, were used to train a classification model on a downstream task. These datasets were only used for evaluation purposes and were not included in training Fairflow.

\begin{enumerate}
\item{\textbf{Bias-in-bios}:}
\sloppy
 This dataset provided by De-Arteaga et al. \cite{de2019bias} contains Wikipedia profiles of professionals. 
 The dataset originally contained labels corresponding to 28 distinct professions alongside the gender labels of the profiled individuals.
 We reclassified the professions into binary labels, aligning them with male-dominated and female-dominated occupations according to gender distribution.
 This categorization was done for two reasons. 
The first was to simplify the classification task from multiclass to binary. Secondly, this enabled us to easily induce bias by creating an imbalance between gender and class labels.

 \item{\textbf{ECHR}:}
The ECHR dataset by Chalkidis et al. \footnote{https://huggingface.co/datasets/coastalcph/fairlex} \cite{chalkidis-etal-2022-fairlex}
 contains case facts from the European Court of Human Rights (ECHR) on human rights breaches by European states.
 It further contains information on the gender of the applicant, human rights articles that were violated, and the defendant state (Central-Eastern European states \textit{v.} all other states). The primary classification task here was to predict the defendant's state based on the case facts.

\item{\textbf{Jigsaw}:}
This dataset\footnote{https://www.kaggle.com/c/jigsaw-unintended-bias-in-toxicity-classification} contains public comments from the now defunct online platform Civil Comments. The primary classification task for this dataset was toxicity detection.

\end{enumerate}

For all the evaluation datasets, we maintained a balanced gender and class label distribution in the test sets as shown in Table~\ref{tab:test_dataset}. The training sets for the Bias-in-bios and the Jigsaw datasets were sampled with an imbalance to induce bias following the observations of Dixon et al. \cite{dixon2018measuring}. The training set for ECHR was left relatively balanced with the additional purpose of providing a baseline.

 \begin{table*}[tbh]
     \centering
     \begin{tabular}{c|c|cccc|ccccc} \hline
           &  & & \multicolumn{3}{c|}{Train} &  & Test &   \\ 
          \textbf{Dataset}&  \textbf{Task} & & \begin{tabular}[c]{@{}c@{}}\textbf{Number}\\ (K) \end{tabular} & \begin{tabular}[c]{@{}c@{}}\textbf{Positive}\\ \textbf{class} \%\end{tabular} & \begin{tabular}[c]{@{}c@{}}\textbf{Females in }\\ \textbf{Pos.} \%\end{tabular}&  \begin{tabular}[c]{@{}c@{}}\textbf{Number}\\ (K) \end{tabular}& \begin{tabular}[c]{@{}c@{}}\textbf{Positive}\\ \textbf{class} \%\end{tabular} & \begin{tabular}[c]{@{}c@{}}\textbf{Females in} \\ \textbf{Pos.} \%\end{tabular} \\ \hline
          \textit{Bias-in-bios}& Career &  & 18 & 50 & 12 &  4 & 50 & 50 \\
          \textit{ECHR} & State &  & 7 & 18 & 41 & 1 & 50 & 50\\
          \textit{Jigsaw} & Toxicity & & 5 & 47 & 77 & 1 & 50 & 50\\ 
     \end{tabular}
     \caption{Evaluation dataset statistics: The test sets are balanced with regard to gender and labels.}
     \label{tab:test_dataset}
 \end{table*}

\subsection{Comparative techniques}

We implemented two variants of FairFLow: \textit{FairFLowV1} and \textit{FairFLowV2}, and compared them to three CDA setups. 
1) \textit{original} is the unaugmented original text;
2) \textit{Hall-M} uses the direct word-substituion approach proposed by Hall-Maudsley et al \cite{hall-maudslay-etal-2019-name};
3) \textit{Hall-M + BART} is a BART model fine-tuned with counterfactuals generated by Hall-Maudsley et al.;
4) \textit{FairFlowv1} is a BART model fine-tuned with our error correction scheme applied to counterfactuals from Hall-Maudsley et al.; it follows the same approach of FairLow in Fig.~\ref{fig: architecture} but with a manually compiled dictionary.
5) \textit{FairFlowv2} is a BART-model fine-tuned with our full approach in Fig.~\ref{fig: architecture}.
We take \textit{Hall-M} and \textit{Hall-M + BART} as our baseline approaches. 
We excluded \textit{perturber} by Qian et al. \cite{qian2022perturbation} from our evaluation since the objective of their approach significantly differs from ours; as elaborated in Section~\ref{sec: related_lit}. 

\section{Evaluation and results}
\label{sec: results}

\begin{figure}
    \centering
    \includegraphics[clip, trim={7cm 2cm 7cm 2cm},width=0.95\linewidth]{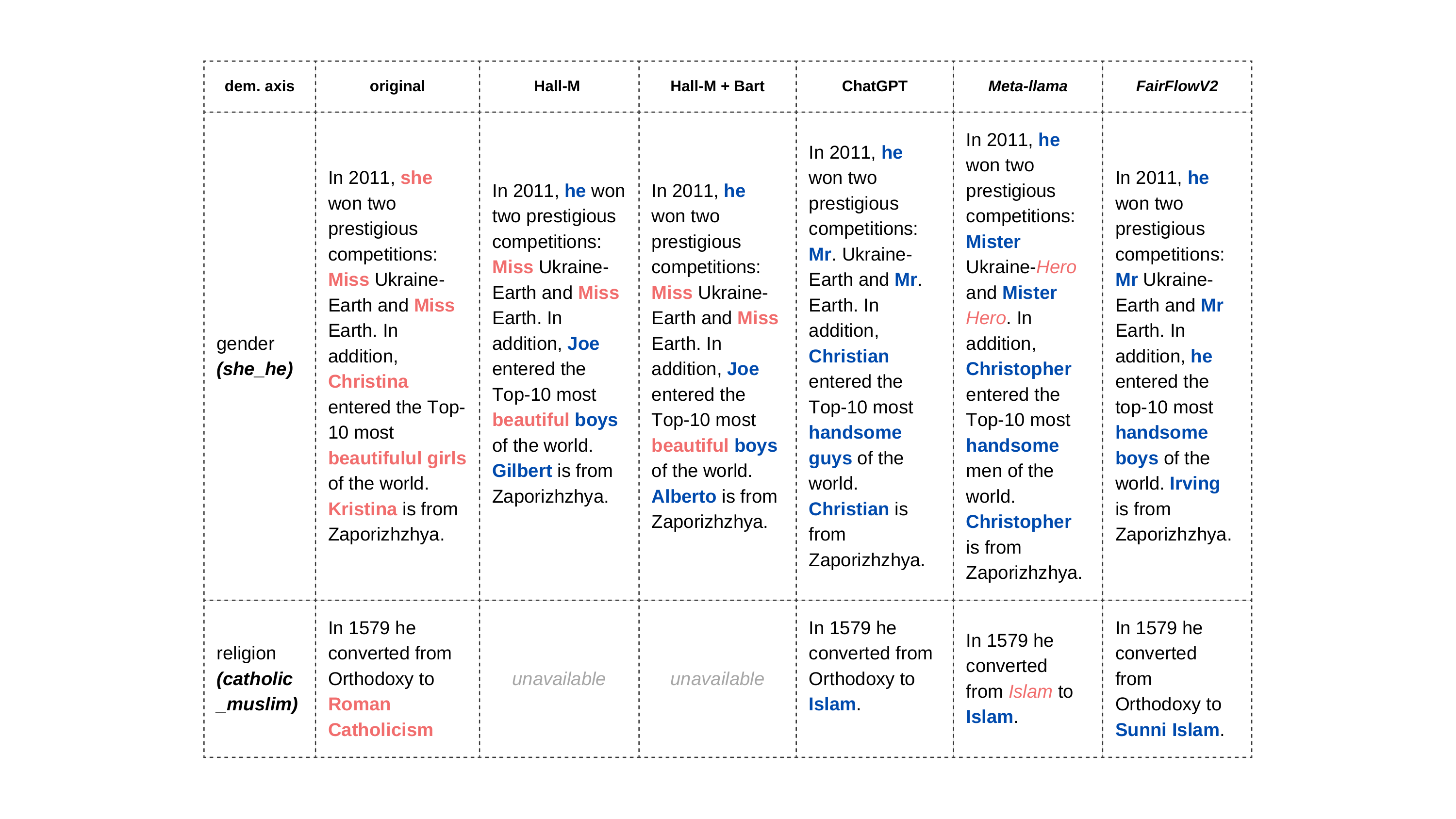}
    \caption{Text samples from Bias-in-bios and Wikipedia demonstrate that \textit{FairFlow} and \textit{ChatGPT-4} generate more robust counterfactual texts. Compared to \textit{ChatGPT-4}, \textit{Meta-llama-3-8B-Instruct} generates more inaccurate counterfactuals.}
    \label{fig: text samples}
\end{figure}

We quantitatively evaluated our approach using three main criteria: \textit{utility}, \textit{extrinsic bias mitigation}, and \textit{task performance}. 

\subsection{Utility}
By utility, we refer to how realistic and effective the generated counterfactuals are by computing their fluency (perplexity) and gender transfer accuracy.

\subsubsection{Grammatical correctness and fluency}
We used a referenceless fluency metric due to the relative unavailability of parallel data. As we noted earlier, the parallel data used by Qian et al. only contains counterfactuals for only specific user-defined entities and is thus not suitable for evaluating our work.
Similar to Wu et al. \cite{wu2021polyjuice}, we score fluency by computing the perplexity of the generated text using pretrained GPT-2 \cite{radford2019language}.
A low perplexity implies that a given text conforms well to the probabilistic distribution of natural text as learned by the pretrained language model.

Based on our earlier assertion about how out-of-context substitutions impair fluency, our error correction approach should expectedly increase fluency (reduce perplexity). We confirm this in Table~\ref{tab: ppl_transfer} as we see that fluency is consistently improved in both \textit{FairFlowV1} and \textit{FairFlowV2}.

\begin{table}[h]
\centering
\resizebox{\linewidth}{!}
{
\begin{tabular}{l|ccc|lll}
\hline
 \multirow{2}{*}{\textbf{Approach}} & \multicolumn{3}{c|}{\textit{PPL} $\downarrow$} &
\multicolumn{3}{c}{\textit{Transfer Accuracy} $\uparrow$ } \\ 
\textbf{} & \textbf{Bios } & \textbf{Jigsaw } & \textbf{ECHR } & {\textbf{Bios } } & {\textbf{\textbf{Jigsaw}}} & \textbf{ECHR }\\
\hline
\textit{Original}* & 41.023 & 69.67 & 32.88 & 0.04 & 15.96 & 36.14 \\
\textit{Hall-M} & 43.51 & 76.37 & 33.70 &  98.60 & \textbf{79.00} & 75.10  \\
\textit{Hall-M + BART} & 47.59 & 83.76 & 39.93 & 98.70 &  78.50 & 71.10\\  
\rowcolor{Gray}\rowcolor{Gray} \textit{FairFlowV1}& 42.77 & 65.80 & 33.70 & \textbf{98.91} & 77.99 &74.69\\
\rowcolor{Gray} \textit{FairFlowV2} & \textbf{39.86} & \textbf{63.99} & \textbf{33.33} & 98.51  &70.736& \textbf{76.51}  \\ 
\end{tabular}
}
\caption{
PPL (\textit{left}) of generated text using various CDA techniques. Lower scores indicate better fluency.
Gender transfer accuracy (\textit{right}) of the various CDA interventions. This indicates the percentage of counterfactual instances that were correctly resolved to new gender styles. \emph{The original samples have very low accuracies because original gender is preserved}.}
\label{tab: ppl_transfer}
\end{table}

\subsubsection{Transfer accuracy}
Here, similar to Tokpo et al. \cite{tokpo-calders-2022-text}, we computed the percentage of texts that were converted from the source attribute to the target attribute, i.e., female to male or vice versa.
We fine-tuned a BERT model to predict the gender of the text.
We quantified gender transfer accuracy as \(1-probability\_of\_original\_attribute\). We expect the original text to have a very low transfer accuracy, as its attributes would remain the same. 
As shown in Table~\ref{tab: ppl_transfer}, FairFlowV2 especially shows strong fluency scores whilst maintaining a good transfer accuracy. 
This shows that automating the dictionary generation process does not materially impair transfer accuracy.

\subsection{Extrinsic bias mitigation}
We trained a BERT classifier using the downstream classification tasks corresponding to the respective datasets and computed the \emph{True Positive rate difference (TPRD)} and \emph{False Positive rate difference (FPRD)} between two gender groups as in the case of De-Arteaga et al. \cite{de2019bias}. 
$TPRD = P(\hat{y}=1|y=1,A=a) - P(\hat{y}=1|y=1,A=a^{\prime})$ and 
$FPRD = P(\hat{y}=1|y=0,A=a) - P(\hat{y}=1|y=0,A=a^{\prime})$.
Where $y$ is the true label, $\hat{y}$ is the predicted label, and $A$ is the gender group variable. 

We show in Table~\ref{tab: bias_mitigation} consistently high TPRD scores for FairFlow1; this further buttresses the evidence that our approach to error correction works effectively and enhances bias mitigation whilst improving fluency. 
Similar to our findings for transfer accuracy, we find that automating dictionary compilation does not compromise bias mitigation much, as FaiFlowV2 maintains a good mitigating effect.

\begin{table}[h]
\centering
\resizebox{\linewidth}{!}
{
\begin{tabular}{l|ccc|ccc}
\hline
 \multirow{2}{*}{\textbf{Approach}} & \multicolumn{3}{c|}{\textit{TPRD} $\downarrow$ } &
\multicolumn{3}{c}{\textit{FPRD} $\downarrow$ } \\ 
\textbf{} & \textbf{Bios} & \textbf{Jigsaw } & \textbf{ECHR } & {\textbf{Bios } } & {\textbf{\textbf{Jigsaw }}} & \textbf{ECHR}\\
\hline
\textit{Original*} & 0.133 & 0.120 & 0.000     &  0.151 &   0.160 &  0.0  \\
\textit{Hall-M}& 0.055 & 0.010 & 0.030 &  0.071 &   0.070 &  0.0 \\
\textit{Hall-M + BART} &    0.051   & 0.025 & 0.010   &   0.074 & \textbf{0.060} &  0.0\\ 
\rowcolor{Gray} \textit{FairFlowV1 } & \textbf{0.044}   & \textbf{0.005} & \textbf{0.000}   & \textbf{0.065} &  0.065 &  0.0 \\
\rowcolor{Gray}\textit{FairFlowV2} & 0.057  & 0.040 & 0.010   & 0.070 &  0.080 &   0.0\\

\end{tabular}

}
\caption{Extrinsic fairness: 
TPRD -- True positive rate difference between male and female text instances.
FPRD -- False positive rate difference between male and female text instances.}
\label{tab: bias_mitigation}
\end{table}

\subsection{Task performance}
We carried out the task performance test to observe the extent to which bias mitigation impacts the task model's performance. Because we maintain a balanced distribution for our test sets, we expect the fairer models to have better performance. Specifically, we computed the accuracy and F1 scores for the default classification task of the respective datasets.
 In Table~\ref{tab: acc_table}, FairFlow1 shows the most improved performance in general, particularly in accuracy. We again show from the strong performance of FairFlowV2, how effective an automatically generated dictionary could be.

\begin{table}[h]
\centering
\resizebox{\linewidth}{!}
{
\begin{tabular}{l|ccc|ccc}
\hline
 \multirow{2}{*}{\textbf{Approach}}& \multicolumn{3}{c|}{ACC $\uparrow$} &
\multicolumn{3}{c}{F1 $\uparrow$} \\ 
\textbf{} & \textbf{Bios} & \textbf{Jigsaw } & \textbf{ECHR } & {\textbf{Bios } } & {\textbf{\textbf{Jigsaw }}} & \textbf{ECHR }\\
\hline
\textit{Original*} & 91.20 & 88.50 & 97.60     &  47.92  &  48.95 &  52.86 \\
\textit{Hall-M}& 92.53 & 90.25 & 97.83 &   48.38  &   48.62 &  52.98 \\
\textit{Hall-M + BART} &  92.64     & 90.62 & 97.36   & \textbf{48.48}  &  49.49  &  52.73\\ 
\rowcolor{Gray} \textit{FairFlowV1 } &  \textbf{92.97}    & \textbf{90.75} & 98.08  &  48.32 &  \textbf{49.69} &  53.11\\
\rowcolor{Gray} \textit{FairFlowV2} & 92.81    & 90.00 & \textbf{98.32}   & 48.36 &  49.21 &   \textbf{53.24}\\

\end{tabular}

}
\caption{Task performance: Accuracy and F1 scores of classification tasks. FairFLow1 shows better performance scores in general. FairFlow2 maintains a significant bias mitigating effect despite an automated dictionary approach.}
\label{tab: acc_table}
\end{table}

\subsection{Qualitative analysis and key observations}
\label{sec: qualitative_analyses}
By analyzing samples from FairFlow, ChatGPT, and the comparative models, we find that FairFLow and ChatGPT have the most grammatically coherent counterfactuals.
Additionally, we find that:
\begin{enumerate}
    \item \textit{\textbf{Automating the dictionary compilation process does not materially impair counterfactual generation.}} As shown in Fig.~\ref{fig: text samples}, even with a dictionary that was automatically compiled, FirFlowV2 generates fluent and plausible counterfactuals. This is aided by the combination of the error correction scheme, which makes it more robust to grammatical errors and helps it generalize better.
    \item \textit{\textbf{A model fine-tuned on erroneous data mimics those errors.}}
    We observe that the error correction approach incorporated in FairFlow makes the model more robust, fluent, and grammatically coherent.
     The direct word replacement technique (\textit{Hall-M}) is unable to replace out-of-dictionary words. The output of \textit{Hall-M + BART} mirrors the same errors as \textit{Hall-M}, showing that a generative model fine-tuned on erroneous data will mimic those errors.
     
    \item \textit{\textbf{ChatGPT generates good counterfactuals but has practical limitations}}. We observe that, in general, ChatGPT generates good counterfactuals in zero-shot settings but is inefficient at generating counterfactuals on a large scale in low-resource environments. It is more costly to deploy in terms of access and infrastructural demands. 
    Secondly, ChatGPT shows inconsistencies in generating counterfactuals for names, as it tends to skip some names for which counterfactuals could have been generated. This is more so if the names refer to public figures, which occasionally leads to grammatical incoherent outputs. 
    This can, however, be addressed by adapting the input prompts and improving instructions through few-shot examples that intuitively describe the setting.
   The manner in which ChatGPT handles names can also be advantageous because it may preserve factuality of the text better, which may be a more desirable attribute in certain contexts.
    We also observed some irregular counterfactuals from \textit{Meta-llama-3-8B-Instruct} in a zero-shot setting, as shown in Fig.~\ref{fig: text samples}. Some of the counterfactuals it generated impacted the original context of the text, which should have been retained.
\end{enumerate}

\section{Conclusion}
In this paper, we highlight some issues that pertain to dictionary-based word-substitution counterfactual data augmentation techniques. We discuss how these techniques, relying on manually compiled dictionaries, are prone to grammatical incoherence and lack generalization outside dictionary terms. We discuss how a model-based approach is primarily inhibited by the relative unavailability of parallel corpora for training. In light of this: 1) we propose an automated dictionary generation approach that can automatically extract and generate word-pairs from a corpus with little human intervention; 2) we propose an error correction approach that can be used to generate fluent and grammatically coherent parallel text to train a generative model for CDA; 3) we combine these approaches to fine-tune a BART model for the purpose of generating counterfactual texts (we call the resulting model \textit{FiarFLow}); 4) we show that our error correction approach significantly improves the fine-tuned model's fluency and bias-mitigating effect; 5) we also show that automating the dictionary compilation process comes at little cost to the performance of the CDA model and is a viable solution in settings where human intervention is challenging.

\section*{Limitations}
\label{sec: limitations}
The primary limitation of our work is the lack of exploration into more diverse demographic and language domains. The work mostly focuses on (binary) gender bias in English, which is a significant limitation, considering how nuanced gender can be in other languages. Due to the relative unavailability of CDA test resources in other demographic domains, such as race, the scope of evaluation in these areas is limited. Our future work will be directed towards addressing these research directions.

Another limitation of this work is its reliance on the tokenization scheme used by the embedding model, which means that words expressed in multiple subtokens are not included in the automatic compilation of the dictionary.

\section*{Ethics Statement}
From an ethical perspective, the primary point to keep in mind regarding the use of counterfactual models is their impact on factuality. Since CDA approaches are designed to be \textit{counterfactual}, they should be used cautiously in sensitive domains where factuality is essential.
Secondly, CDA bias mitigation techniques like FairFlow do not automatically guarantee fairness; hence, they must be used with that understanding.

\section*{Acknowledgements}
Ewoenam Kwaku Tokpo received funding from the Flemish Government under the ``Vlaams AI-Onderzoeksprogramma'' (Flanders AI Research Program).
We also thank Marco Favier for sharing his insights and engaging in valuable discussions.

\bibliographystyle{splncs04}
\bibliography{custom}

\end{document}